\documentclass{article}
\usepackage{pifont}
\usepackage{iclr2026_conference,times}

\usepackage{amsmath,amsfonts,bm}

\def\eqref#1{equation~\ref{#1}}

\def\1{\bm{1}}

\DeclareMathAlphabet{\mathsfit}{\encodingdefault}{\sfdefault}{m}{sl}
\SetMathAlphabet{\mathsfit}{bold}{\encodingdefault}{\sfdefault}{bx}{n}

\usepackage{hyperref}
\usepackage{etoc}
\usepackage[toc,page]{appendix}
\usepackage{url}
\usepackage{arydshln}
\usepackage{rotating} 
\usepackage{booktabs} 
\usepackage{multirow} 
\usepackage{array} 
\usepackage{amsmath} 
\usepackage{xcolor}
\usepackage{siunitx}
\usepackage{epigraph}
\usepackage[table]{xcolor}
\usepackage{makecell}
\usepackage{bbding}
\usepackage{tcolorbox} 
\usepackage{graphicx} 

\definecolor{SingleImg}{RGB}{217,237,247}  
\definecolor{MultiImgOpen}{RGB}{223,240,216}
\definecolor{MultiImgProp}{RGB}{241,233,244}
\definecolor{OursColor}{RGB}{252,228,225}  

\title{Can World Models Benefit VLMs for World Dynamics?}

\author{Kevin Zhang\textsuperscript{1,*},
Kuangzhi Ge\textsuperscript{1,*}, 
Xiaowei Chi\textsuperscript{2,*}, 
Renrui Zhang\textsuperscript{3}, 
Shaojun Shi\textsuperscript{1}, \\
\textbf{Zhen Dong\textsuperscript{4},
Sirui Han\textsuperscript{2,\Envelope}, 
Shanghang Zhang\textsuperscript{1,\Envelope}} \\ 
\small\textsuperscript{1} Peking University 
\small\textsuperscript{2} Hong Kong University of Science and Technology \\ 
\small\textsuperscript{3} Chinese University of Hong Kong 
\small\textsuperscript{4} University of California, Santa Barbara \\
* Equal contribution, \Envelope$ $ Corresponding author
}

\definecolor{singleopensource}{HTML}{E0F2F1} 
\definecolor{multiopensource}{HTML}{E8F5E9}  
\definecolor{multiproprietary}{HTML}{FFFDE7} 
\definecolor{ours}{HTML}{FFEBEE}             

\definecolor{firstplace}{HTML}{D32F2F}  
\definecolor{secondplace}{HTML}{1976D2} 
\definecolor{thirdplace}{HTML}{4CAF50}  

\definecolor{softnote}{RGB}{245,245,255}
\definecolor{strongnote}{RGB}{160,160,255}

\iclrfinalcopy 

\begin{document}

\maketitle

\begin{abstract}

Trained on internet-scale video data, generative world models are increasingly recognized as powerful world simulators that can generate consistent and plausible dynamics over structure, motion, and physics.
This raises a natural question: \textbf{\textit{with the advent of strong video foundational models, might they supplant conventional vision encoder paradigms for general-purpose multimodal understanding?}}
While recent studies have begun to explore the potential of world models on common vision tasks, these explorations typically lack a systematic investigation of generic, multimodal tasks. 
In this work, we strive to investigate the capabilities when world model priors are transferred into Vision-Language Models (VLMs): we re-purpose a video diffusion model as a \emph{generative encoder} to perform a single denoising step and treat the resulting latents as a set of visual embedding.
We empirically investigate this class of models, which we refer to as World-Language Models (WorldLMs), and we find that generative encoders can capture latents useful for downstream understanding that show distinctions from conventional encoders. 
Naming our best-performing variant \textbf{Dy}namic \textbf{V}ision \textbf{A}ligner (\textbf{DyVA}), we further discover that this method significantly enhances spatial reasoning abilities and enables single-image models to perform multi-frame reasoning.
Through the curation of a suite of visual reasoning tasks, we find DyVA to surpass both open-source and proprietary baselines, 
achieving state-of-the-art or comparable performance. 
We attribute these gains to WorldLM's inherited motion-consistency internalization from video pre-training.
Finally, we systematically explore extensive model designs to highlight promising directions for future work. 
We hope our study can pave the way for a new family of VLMs that leverage priors from world models and are on a promising path towards generalist vision learners.

Project page: \href{https://dyva-worldlm.github.io/}{https://dyva-worldlm.github.io/}.
\end{abstract}

\begin{figure}[h]
    \centering
    \includegraphics[width=0.95\linewidth]{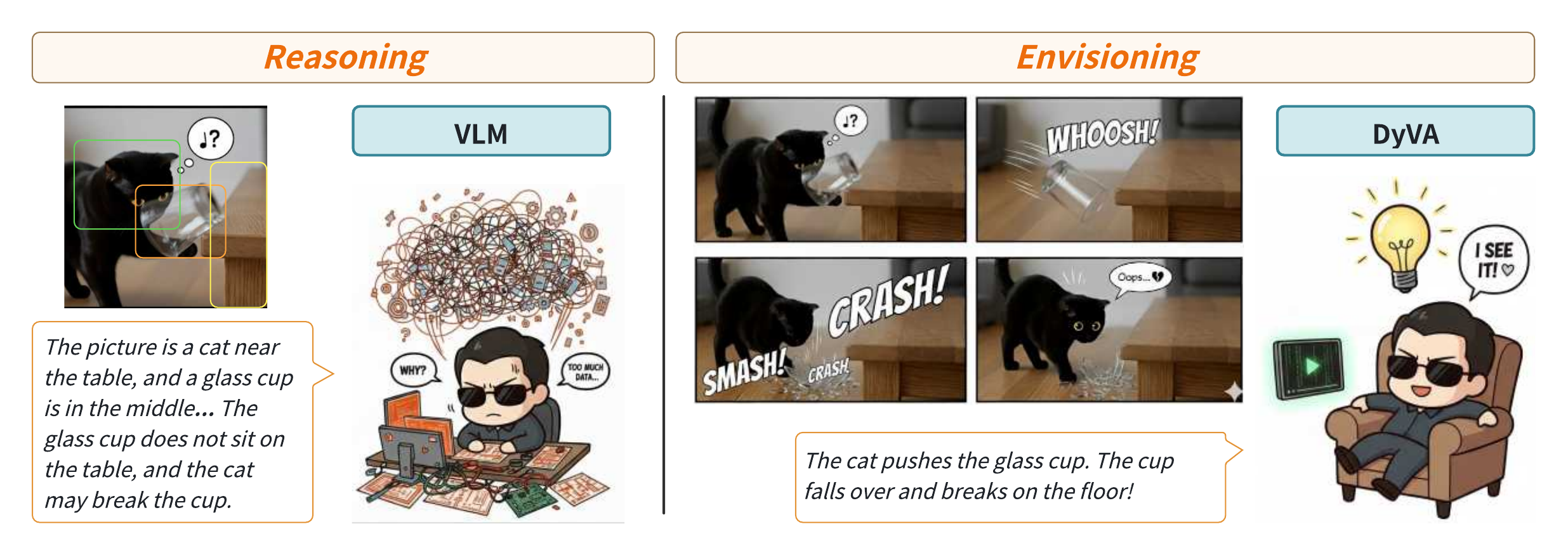}
    \caption{\textbf{What will happen?} From reasoning to dynamic intuition — comparing how VLM and WorldLM understand and predict real-world events. }
    \label{fig:main}
\end{figure}

\begin{figure}[htbp]
    \centering
    \includegraphics[width=\linewidth]{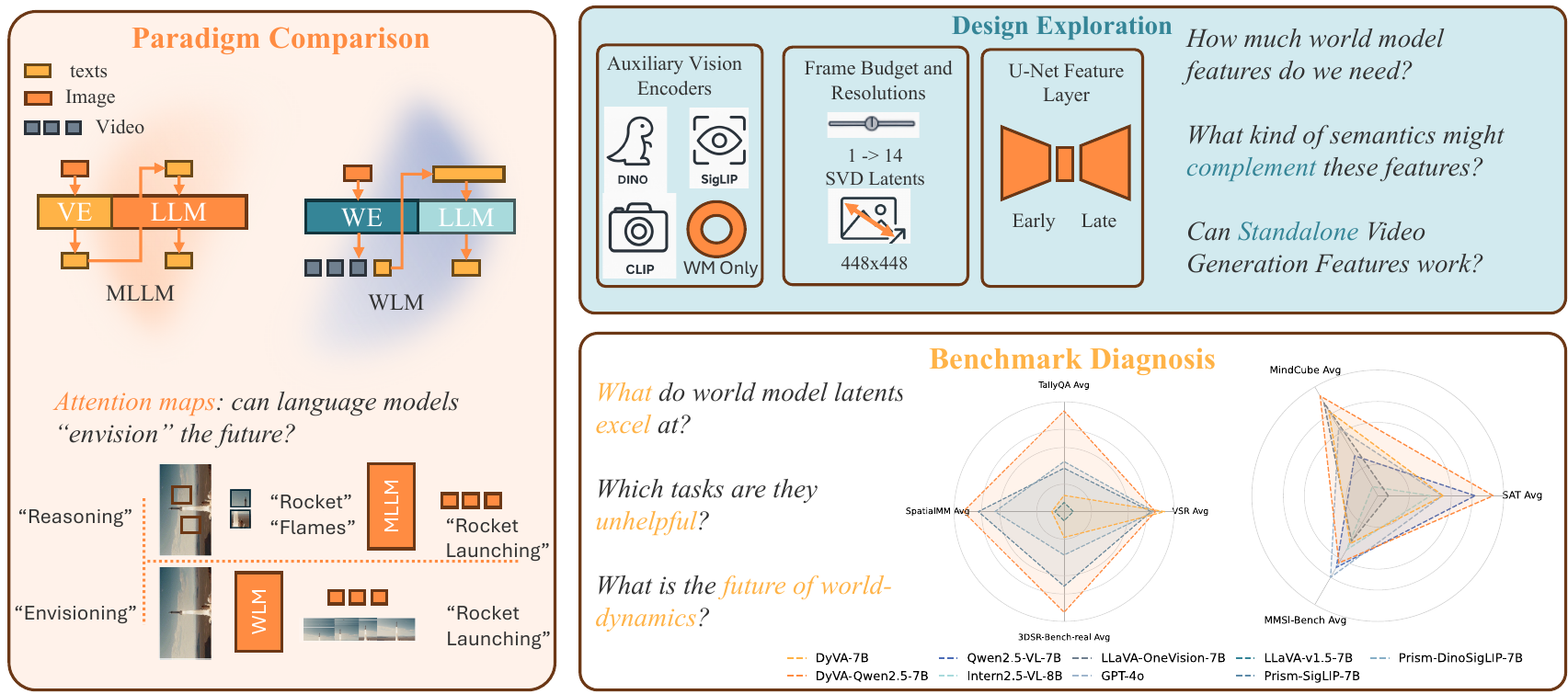}

    \caption{Our analysis is structured around three dimensions: (i) Paradigm comparison between static and generative encoders (e.g., SigLIP vs. SVD); (ii) Benchmark diagnosis, revealing world model latents' strength (e.g., spatial/multi-frame reasoning) and weaknesses (e.g., language-heavy tasks); and (iii) Design-space exploration, probing different auxiliary encoders, resolutions, and training recipes to understand how world-model features aid visual understanding. }
    \label{fig:analysis-dashboard}
\end{figure}

\section{Introduction}

World models, originally proposed in cognitive science to explain how humans predict and plan in their environments~\citep{tolman1948cognitive}, have recently emerged as powerful tools in machine learning. 
Generative world models, such as video generation models (VGMs)~\citep{agarwal2025cosmospredict1,openai2024sora,wan2025wan,hu2023gaia1generativeworldmodel,Blattmann2023Stable,yang2025cogvideox,guo2023point,guo2025can,chen2025mint} that are trained on internet-scale video data, encode strong priors over objects, spatial layouts, and dynamics. These priors allow them to predict plausible future scenarios that are consistent in 3D structure and physically coherent in motion.

However, a largely overlooked implication of World Models is that the ability to generate coherent futures signals a form of semantic understanding of visual dynamics; this difference between visual generation and understanding has shaped a decade of representation learning.
This suggests that world models can be more than generators—they may serve as transferable encoders that enrich downstream tasks with spatial, temporal, and predictive signals. As a result, recent work has attempted to use video generation backbones for visual perception tasks~\citep{acuaviva2025generationgeneralizationemergentfewshot, wiedemer2025videomodelszeroshotlearners}.

In this work, we ask a foundational question:
\textit{\textbf{can generative models surpass current vision understanding paradigms for generic, multimodal understanding?}}

To empirically investigate the current capabilities of video generation models, we introduce a simple yet effective framework applying them to Vision–Language Models (VLMs). We specifically explore this by evaluating the applicability of predictive world models on a generic multimodal task - Visual Question Answering (VQA) — to assess their broader potential as \textbf{generalizable vision encoders}.
Currently, mainstream VLMs primarily rely on 
ViT-based encoders such as CLIP~\citep{Radford2021CLIP}, SigLIP~\citep{zhai2023sigmoidlosslanguageimage}, and DINO~\citep{caron2021emergingpropertiesselfsupervisedvision,oquab2024dinov2learningrobustvisual}, which extract visual semantics from image patches and are then projected as visual tokens into language backbones.
While these encoders are semantically aligned, they are limited by temporal reasoning and weaken spatial grounding when multiple views or sequential cues are present.
On the other hand, we re-purpose a world model (Stable Video Diffusion, or SVD) as a \textbf{Generative Encoder}. Our core mechanism is to extract latent features from a \textbf{single denoising step} of its U-Net. This single step, we hypothesize, captures the low-dimensional world-dynamics prior sufficient for downstream understanding. These dynamics-aware latents are then fused with static image features (e.g., SigLIP) and projected into the Large Language Model (LLM). The design is very efficient: all encoders remain frozen, with only lightweight projectors and the LLM being trained.

To this end, we conduct a systematic investigation comparatively evaluating this class of models, which we refer to as World-Language Models (WorldLMs). Our findings are as follows:

\begin{itemize}
    \item \textbf{Shift in Reasoning Paradigm.} The generative prior alters the model's reasoning process. It moves beyond describing static content to envisioning dynamic possibilities.
    \item \textbf{Zero-shot Multi-Frame Adaptation.} Trained only on single images, the generative encoder enables emergent multi-frame reasoning without multi-image training.
    On multi-frame visual reasoning, DyVA achieves state-of-the-art or comparable performance with flagship models such as Qwen2.5-VL~\citep{bai2025qwen25vltechnicalreport} and GPT-4o~\citep{openai2024gpt4ocard}.
    \item \textbf{We empirically identify the regimes where video priors help.} Our ablations and diagnostics separate the settings in which SVD latents strengthen spatial reasoning from those where they dilute semantic grounding, guiding future designs.
\end{itemize}

Our best-performing WorldLM variant, \textbf{Dy}namic \textbf{V}ision \textbf{A}ligner (\textbf{DyVA}), exemplifies this paradigm shift. In zero-shot evaluations on challenging multi-frame reasoning benchmarks, DyVA decisively surpasses even proprietary models, for instance, a \textbf{28.3\%} lead on the \textbf{MindCube} benchmark over the GPT-4o model. This provides strong evidence that the ability to predict is a powerful, perhaps essential, foundation for stronger representation learning.

As shown in Figure~\ref{fig:analysis-dashboard}, we systemically organize our investigation revolving around three pillars:

\textbf{Paradigm comparison.} World-model encoders versus static encoders reveal distinct strengths: world-model latents benefit spatial and multi-frame reasoning, while static encoders excel on semantics-heavy benchmarks.

\textbf{Benchmark diagnostics.} 
Through curated evaluation sets including MindCube~\cite{yin2025mindcube}, SAT-Bench~\cite{ray2024sat}, VSR~\cite{vsr}, we find that DyVA surpasses both open-source and proprietary baselines on out-of-domain tasks, achieving \textbf{state-of-the-art performance on MindCube}.
Given that SVD is pre-trained on temporally coherent video–text pairs, we show that dynamics-aware latents particularly boost object relations, cross-view understanding, and multi-frame spatial reasoning, while offering less gain on tasks requiring stronger language priors.

\textbf{Design-space exploration.} We analyze different encoder setups to identify when predicted latents help or hinder performance and analyzing the co-training of U-Net and VAE layers with text loss, laying the groundwork for a new class of WorldLMs exploiting world-model priors.
\begin{figure}[htbp]
    \centering
    \includegraphics[width=\linewidth]{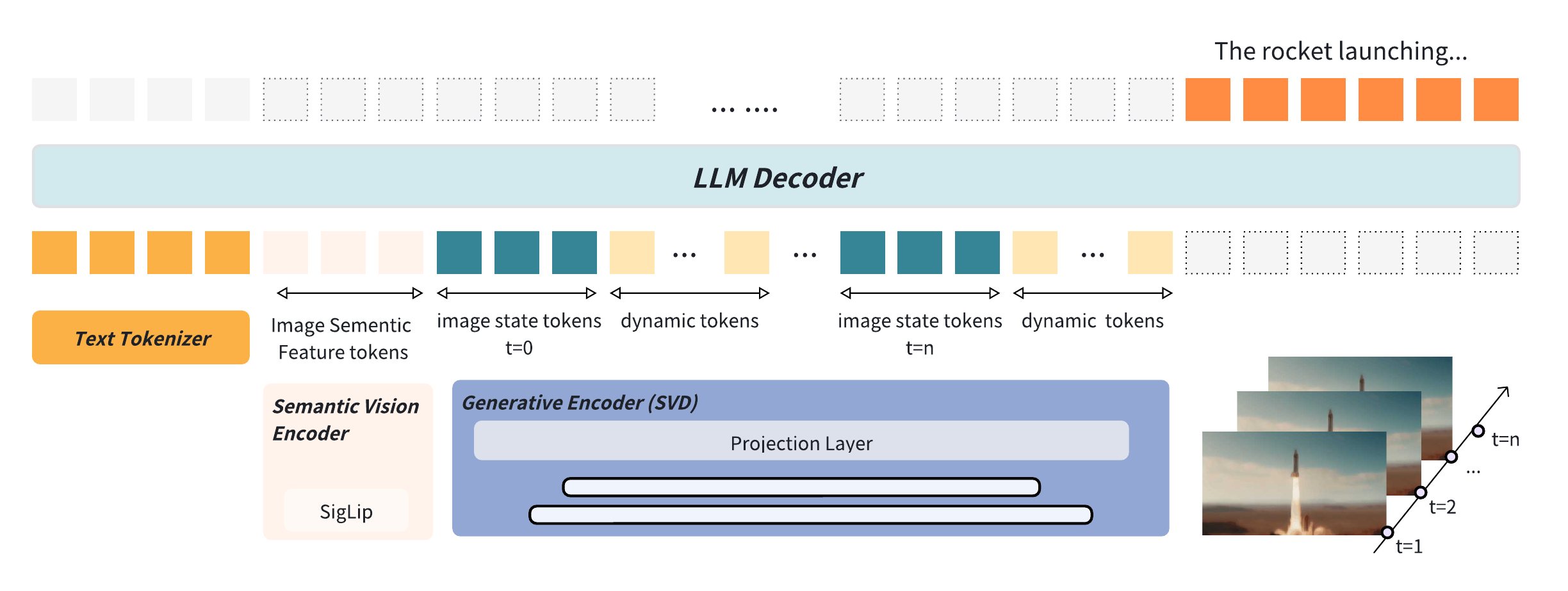}
    \caption{\textbf{WorldLM Pipeline.} A SigLIP encoder extracts semantic features from the input image. Concurrently, a Generative Encoder generates dynamic state tokens to capture temporal changes, using evenly spaced keyframe slots. All visual tokens are projected into a shared embedding space, concatenated with text tokens, and then fed into the LLM decoder.}
    \label{fig:pipeline}
\end{figure}
\section{Preliminary}
\label{sec:preliminary}
We first lay a groundwork for our analysis with: 1) a framework to incorporate the dynamic features of a world model into a multimodal language model (which we term \textbf{WorldLM}), 2) a training recipe, and 3) an implementation of inference supporting both single and multi-image datasets.

\paragraph{Framework.} Given an input image $x_{img} \in \mathbb{R}^{H \times W \times C}$ and a text prompt $u_{prompt}$, traditional VLMs such as LLaVA~\citep{liu2024improvedbaselinesvisualinstruction}, QwenVL~\citep{bai2025qwen25vltechnicalreport}, InternVL~\citep{internvl25}, DeepSeekVL~\citep{lu2024deepseekvl}, and Prismatic-VLMs~\citep{karamcheti2024prismaticvlmsinvestigatingdesign}, process the input with an architecture consisting of three core components:

\begin{itemize}
    \item \textbf{Semantic Vision Encoder.} $x_{img}$ is processed by a frozen pre-trained ViT-based~\citep{dosovitskiy2021imageworth16x16words} encoder $V_\omega$, e.g., SigLIP~\citep{zhai2023sigmoidlosslanguageimage}, to extract a sequence of feature embeddings $p_{img} = V_\omega(x_{img})$, where $p_{img} \in \mathbb{R}^{L \times d_{vision}}$, where L is the token length and $d_{vision}$ refers to the vision feature dimension.
    \item \textbf{Projector.} The visual features $p_{img}$ are subsequently mapped into the language model's embedding space by a projector $F_\psi$. This yields a sequence of embeddings $e_{img} = F_\psi(p_{img})$, where $e_{img} \in \mathbb{R}^{L \times d_{text}}$, where $d_text$ is the text feature dimension. The projector is typically implemented as a simple MLP with GELU activations~\citep{hendrycks2023gaussianerrorlinearunits}.
    \item \textbf{LLM Backbone.} Finally, the language model $\text{LM}_\theta$ autoregressively generates the textual output $u_{out}$. It is conditioned on the concatenated sequence of the projected image features $e_{img}$ and the text prompt embeddings $e_{prompt}$: $u_{out} = \text{LM}_\theta([e_{img}; e_{prompt}])$
\end{itemize}

On the other hand, in WorldLMs, we employ a Generative Encoder to extract dynamic visual information and motion priors of the input image: 

\begin{itemize}
    \item \textbf{Generative Encoder.} We utilize Stable Video Diffusion (SVD)~\citep{Blattmann2023Stable} as our encoder. SVD consists of a VAE~\citep{kingma2022autoencodingvariationalbayes} encoder $\phi$ and a U-Net~\citep{ronneberger2015unetconvolutionalnetworksbiomedical} denoiser $f_\theta$. The input image $x_{img}$ is first embedded by VAE into the latent $z_0$, which is then replicated $T$ times to form the initial video latent $Z_0$. A single Euler integration step~\citep{karras2022elucidatingdesignspacediffusionbased} is then applied to yield an updated latent $Z_{1} = Z_{0} + \Delta\sigma \, f_\theta(Z_{0}, \sigma_0, c)$. Rather than rendering video frames, the final output $D_{img} = \mathrm{Hidden}^{\text{mid}}(f_\theta, Z_{1})$ is extracted from U-Net's middle layers.
\end{itemize}

As shown in Fig.~\ref{fig:pipeline}, semantic features $p_{img}$ and dynamic features $\tilde H$ are projected by two separate projectors $P_{\text{sem}}$ and $P_{\text{dyn}}$ into the LLM space, yielding $V_s = P_{\text{sem}}(p_{img}) \in \mathbb{R}^{L_s \times d}$ and $V_d = P_{\text{dyn}}(\tilde H) \in \mathbb{R}^{L_d \times d}$. The fused sequence is $V = [V_s; V_d] \in \mathbb{R}^{(L_s+L_d)\times d}$, which, together with prompt embeddings $E_{prompt}$, is fed into the LLM backbone to autoregressively generate answer tokens $u_{out} = \mathrm{LM}_\theta([V;E_{prompt}])$. By fusing both streams, our WorldLM leverages static semantics (from SigLIP) and dynamics-aware priors (from SVD) for multimodal reasoning. 

\textbf{Training recipe.} We adopt the training strategy from Prismatic-VLMs~\citep{karamcheti2024prismaticvlmsinvestigatingdesign}, using single-stage training to align modalities and incorporate dynamic features:
We jointly train both the projectors and the LLM on a mixture of multimodal instruction datasets from LLaVA-1.5~\citep{liu2023visualinstructiontuning}, together with examples from established vision-language benchmarks (e.g., GQA~\citep{hudson2019gqa}, TextCaps~\citep{sidorov2020textcapsdatasetimagecaptioning}), and language-only samples from ShareGPT~\citep{sharegpt}. This training paradigm not only effectively aligns the representations of the generative encoder with the semantic space of the LLM but also improves its compositional generalization, allowing it to reason over both priors of motion and the static features. Remarkably, the entire training process completes in only 10.3 hours on 16$\times$A800 GPUs (\(\approx\)165 GPU-hours) while achieving competitive performance, underscoring the efficiency of our approach.

\paragraph{Inference Protocol} During inference, we employ SigLIP-so400m-patch14-224 as the semantic vision encoder and SVD as the generative encoder with an image resolution of $448 \times 448$. Shown in Fig.~\ref{fig:pipeline}, for $K$ input images, we allocate key frames using evenly spaced indices within the $T$-frame latent tensor, replacing the corresponding slots with encoded keyframes before the Euler step, and reuse the resulting latents as visual tokens. For the semantic vision encoder, only the first input image is encoded and concatenated with the input of the generative encoder. Unless otherwise specified, the number of frames ($T$) is set to 8 for both single-image and multi-image inputs.

Following the proposed framework, training setup, and inference principles, we train a family of WorldLM models and designate the optimal ones in \textbf{Dy}namic \textbf{V}ision \textbf{A}lignment as \textbf{\textit{DyVA}}.


\section{Paradigm Comparison}
\label{sec:paradigm}

\textbf{Do WorldLM Encoders Entail Visual Semantics Understanding?}

In this section, we explore how world model latents can benefit visual understanding by contrasting two differentiating encoder paradigms: (i) conventional static encoders such as CLIP and SigLIP that prioritize multimodal semantic alignment, and (ii) WorldLM encoders based on video generation models that generate dynamics-aware latents. 
%
We begin by comparing the most intuitive design to test if WorldLMs can work, by directly replacing the CLIP vision encoder of LLaVA 1.5~\citep{liu2024improvedbaselinesvisualinstruction} with a Generative Encoder (e.g., SVD) following the WorldLM pipeline settings in Fig.~\ref{fig:pipeline}.

\begin{figure}[ht]
    \centering
    \includegraphics[width=\linewidth]{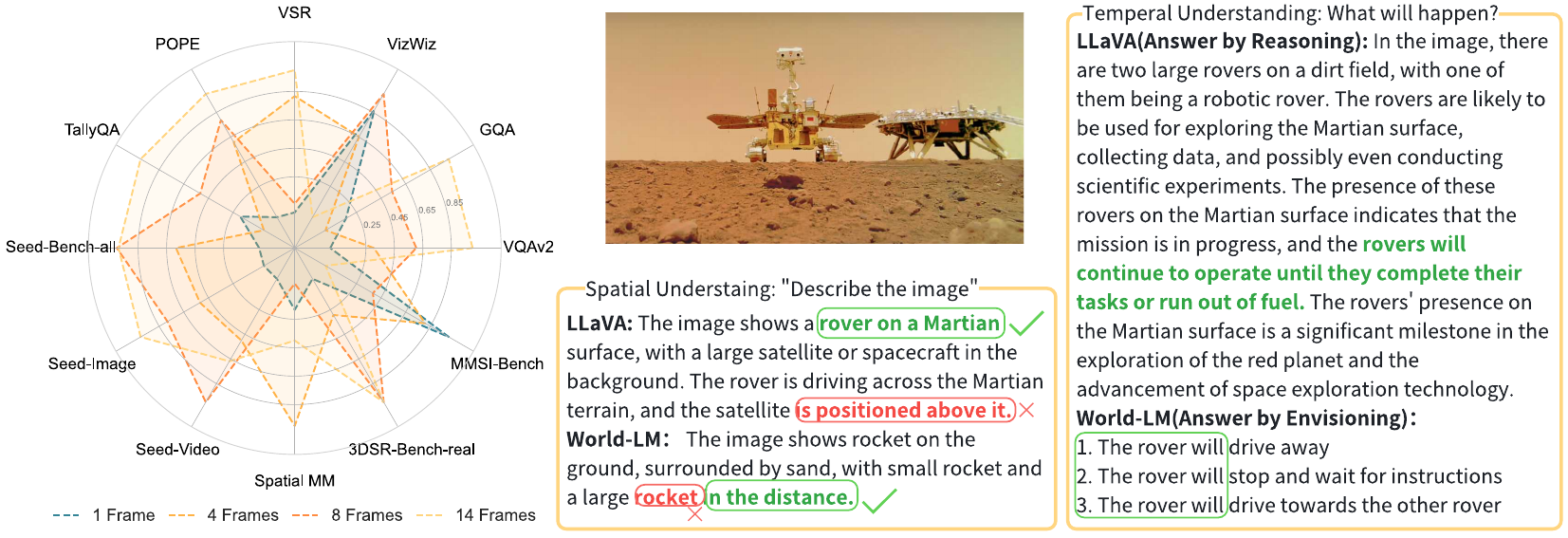}
\caption{\textbf{Paradigm Comparison.} We evaluate predicting 1, 4, 8, and 14 frames with a straightforward WorldLM setup. The radar chart (left) demonstrates that more frames boosts performance across various tasks, especially in visual reasoning. The qualitative example (right) illustrates that our WorldLM exhibits a distinct reasoning paradigm by envisioning, offering concise descriptions, stronger spatial grounding, and more structured temporal foresight compared to LLaVA.}

    \label{fig:paradigm}
\end{figure}

\textbf{Generative encoders exhibit fundamentally different performance.}

We begin with a motivating case study, illustrated in Fig.~\ref{fig:paradigm}. Models leveraging static encoders, such as LLaVA, adopt a \textit{reasoning} paradigm.
The output of LLaVA tends to be more descriptive, describing in depth the details of the given image input. 
WorldLM, on the other hand, employs an \textit{envisioning} paradigm, which not only encodes the current state of the image, but it also performs a prediction of plausible future conditions (e.g., ``will drive away'', and ``drive to the other rover''). 
This case reveals an intrinsic difference between the two paradigms: VLM reasons by the given image's embeddings, whereas WorldLM attends to depict the embeddings of generated predictions. 

\textbf{Multi-frame capture more useful semantic features than Single Frame.} The quantitative~\ref{fig:paradigm} comparison between using different numbers of generated dynamic latents shows its effect on downstream tasks. When the generated frames of the video prediction model increase from 1 to 14, we see a general rising trend of performance on all tasks. 

\textbf{Meanwhile, the vanilla WorldLM performs great on spatial-reasoning tasks.}
Notably, the gains are most pronounced on benchmarks requiring visual reasoning through space and time, such as \textit{SeedBench}, \textit{VSR}, and \textit{TallyQA}. This demonstrates the potential of using world models as dynamics-aware encoders to allow VLMs gain a deeper and more grounded level of spatial understanding.

\textbf{Limitations of WorldLMs.}
Despite the clear advantages in spatial reasoning, our empirical study reveals a critical trade-off in Fig.~\ref{fig:paradigm}, where its performance relative to LLaVA is much lower across all tasks. The case study in Fig.~\ref{fig:paradigm} offers a qualitative explanation for this phenomenon. While our world model correctly grounds the spatial structure of the scene (e.g., ``rocket on the ground... large rocket in the distance''), it hallucinates the semantic identity of the objects, misidentifying the Mars lander and rover as ``rockets''. Therefore, we believe that using a world model as an encoder has the potential to enhance predictive and spatial reasoning tasks, but requires further improvement to ensure basic semantic capabilities.

\section{Benchmark Analysis: Investigation}
\label{sec:benchmark}

\subsection{Experimental Setup}
\label{sec:setup}

We document the configurations, datasets, and training protocols underlying our study. 
Unless otherwise noted, all settings use a 7B-parameter LLaMA-2 LLM backbone, with both SigLIP and SVD encoders frozen during a single-stage instruction tuning. 
Training updates are restricted to lightweight projection layers and the language backbone.

\subsection{Datasets and Evaluation Targets}

Benchmarks vary widely in their emphasis on \emph{spatial grounding}, \emph{temporal coherence}, and \emph{semantic understanding}. 
To assess these dimensions, we curate a suite of open-source \textbf{out-of-domain (OOD)} datasets on which our models have not been trained. 
This allows us to isolate the transferability of world-model priors.

\textbf{Single-image spatial reasoning.}  
We evaluate on benchmarks that probe relational and spatial understanding without temporal context, including    
VSR~\citep{vsr},  
TallyQA~\citep{acharya2018tallyqaansweringcomplexcounting},  
SpatialMM-Obj~\citep{spatialmm},  
and 3DSR-Bench-real~\citep{ma20253dsrbenchcomprehensive3dspatial}.  
Baselines include LLaVA-1.5~\citep{liu2024improvedbaselinesvisualinstruction},  
Prism-SigLIP-7B~\citep{karamcheti2024prismaticvlmsinvestigatingdesign},  
and Prism-DinoSigLIP-7B~\citep{karamcheti2024prismaticvlmsinvestigatingdesign}.

\textbf{Multi-image and temporal reasoning.}  
To assess robustness to sequential inputs and temporal structure, we use MMSI-Bench~\citep{yang2025mmsibench}, SAT-Synthetic~\citep{ray2024sat}, and MindCube~\citep{yin2025mindcube}.  
These benchmarks require models to integrate cues across frames or viewpoints, testing whether world-model latents can enable multi-frame reasoning.  
We compare against both open-source and proprietary large-scale VLMs, including  
Qwen-2.5-VL-7B~\citep{bai2025qwen25vltechnicalreport},  
InternVL-2.5-7B~\citep{internvl25},  
LLaVA-OneVision-7B~\citep{li2024llavaonevisioneasyvisualtask}, 
and GPT-4o~\citep{openai2024gpt4ocard}. Note that all of the compared benchmarks are trained with multi-frame or video data, whereas we train on single images only.

\subsection{Experimental Analysis and Insights}

\begin{table*}[h]
\centering
\caption{
    Performance comparison between DyVA and state-of-the-art methods on multi-image benchmarks SAT Synthetic, MMSI-Bench, and MindCube. DyVA outperforms baselines in these OOD tasks without training on multi-image datasets. The highest average values are in bold. 
}
\label{tab:sat_mmsi_mindcube}

\resizebox{\textwidth}{!}{
\begin{tabular}{@{}l ccccc c ccc c@{}}
\toprule
\multirow{2}{*}{\textbf{Model}} & \multicolumn{6}{c}{\textbf{SAT Synthetic}} & \multicolumn{4}{c}{\textbf{MindCube}} \\
\cmidrule(lr){2-7} \cmidrule(lr){8-11}
& \textbf{Obj Move.} & \textbf{Act. Seq.} & \textbf{Act. Cons.} & \textbf{Goal Aim} & \textbf{Persp.} & \textbf{Avg.} & \textbf{Rot.} & \textbf{Among} & \textbf{Around} & \textbf{Avg.} \\
\midrule
Qwen2.5-VL-7B & 79.29 & 84.70 & 47.83 & 25.84 & 35.17 & 53.16 & 38.76 & 29.50 & 21.35 & 29.26 \\
Intern2.5-VL-8B & 77.74 & 55.49 & 53.74 & 15.03 & 32.61 & 48.06 & 18.68 & 36.45 & 18.20 & 18.68 \\
LLaVA-OneVision-7B & 71.10 & 21.64 & 49.85 & 31.76 & 35.43 & 43.24 & 36.45 & 48.42 & 44.09 & 47.43 \\
GPT-4o & 61.50 & 33.20 & 47.60 & 67.50 & 37.50 & 49.40 & 40.17 & 29.16 & 38.81 & 38.81 \\
\midrule
\rowcolor{blue!10}
\textbf{DyVA-7B} & 49.15 & 57.81 & 49.25 & 53.38 & 40.44 & 49.51 & 37.70 & 43.10 & 49.00 & 44.62 \\
\rowcolor{blue!10}
\textbf{DyVA-Qwen2.5-7B} & 78.83 & 62.13 & 49.85 & 51.86 & 41.72 & \bfseries 55.24 & 37.20 & 39.10 & 51.70 & \bfseries 49.80 \\
\bottomrule
\end{tabular}
}

\resizebox{\textwidth}{!}{
\begin{tabular}{@{}l cccccc cc cc c c@{}}
\toprule
\multirow{3}{*}{\textbf{Model}} & \multicolumn{12}{c}{\textbf{MMSI-Bench}}  \\
\cmidrule(lr){2-13}
& \multicolumn{6}{c}{\textbf{Positional Relationship}} & \multicolumn{2}{c}{\textbf{Attribute}} & \multicolumn{2}{c}{\textbf{Motion}} & \multirow{2}{*}{\textbf{MSR}} & \multirow{2}{*}{\textbf{Avg.}} \\
\cmidrule(lr){2-7} \cmidrule(lr){8-9} \cmidrule(lr){10-11}
& Cam–Cam & Obj–Obj & Reg–Reg & Cam–Obj & Obj–Reg & Cam–Reg & Means & Appr & Cam & Obj & & \\
\midrule
Qwen2.5-VL-7B & 32.3 & 27.7 & 29.6 & 32.6 & 24.7 & 32.5 & 26.6 & 27.3 & 16.2 & 31.6 & 30.3 & 28.70 \\
Intern2.5-VL-8B & 24.7 & 24.5 & 24.7 & 25.6 & 29.4 & 26.5 & 25.0 & 18.2 & 20.3 & 39.5 & 25.8 & 25.90 \\
LLaVA-OneVision-7B & 20.4 & 33.0 & 29.6 & 29.1 & 25.9 & 30.1 & 29.7 & 25.8 & 18.9 & 34.2 & 11.6 & 24.50 \\
GPT-4o & 34.4 & 24.5 & 23.5 & 19.8 & 37.6 & 27.7 & 32.8 & 31.8 & 35.1 & 36.8 & 30.8 & \textbf{30.30} \\
\midrule
\rowcolor{blue!10}
\textbf{DyVA-7B} & 21.5 & 30.9 & 25.9 & 31.4 & 27.1 & 20.5 & 35.9 & 24.2 & 13.5 & 19.7 & 24.2 & 24.90 \\
\rowcolor{blue!10}
\textbf{DyVA-Qwen2.5-7B} & 15.1 & 33.0 & 25.9 & 33.7 & 35.3 & 30.1 & 32.8 & 25.8 & 17.6 & 27.6 & 29.3 & 28.00 \\
\bottomrule
\end{tabular}
}
\end{table*}

\begin{table*}[t]
\centering
\caption{
    Performance comparison of DyVA variants against baselines on various single-image spatial reasoning benchmarks, including VSR, TallyQA, SpatialMM-Obj, and 3DSR-Bench-real. 
    These are Out-of-Domain tasks where models are not trained and perform zero-shot inference. Our results surpass all baseline models. Highest values are highlighted in bold.
}
\label{tab:single_exp}

\resizebox{\textwidth}{!}{
\begin{tabular}{@{}l c ccccccc c@{}}
\toprule
\multirow{2}{*}{\textbf{Models}} & 
\multirow{2}{*}{\textbf{Data}} & 
\multicolumn{8}{c}{\textbf{VSR}} \\ 
\cmidrule(lr){3-10}
& & {\textbf{Topo.}} & {\textbf{Prox.}} & {\textbf{Proj.}} & {\textbf{Direc.}} & {\textbf{Adj.}} & {\textbf{Orien.}} & {\textbf{Unall.}} & {\textbf{Avg.}} \\
\midrule
LLaVA-v1.5-7B & 558k+665k & 52.24 & 50.00 & 54.77 & 50.00 & 50.86 & 48.98 & 57.50 & 52.94 \\
Prism-SigLIP-7B & 665k & 67.48 & 62.50 & 65.63 & 66.67 & 55.17 & 55.10 & 67.50 & 64.97 \\
Prism-DinoSigLIP-7B & 665k & 71.34 & 59.38 & 65.63 & 64.29 & 53.45 & 48.98 & 52.50 & 65.46 \\
\midrule
\rowcolor{blue!10}
\textbf{DyVA-7B} & 665k & 68.90 & 68.75 & 66.74 & 66.67 & 66.38 & 61.22 & 57.50 & \bfseries 67.10 \\
\rowcolor{blue!10}
\textbf{DyVA-Qwen2.5-7B} & 665k & 66.67 & 71.88 & 68.74 & 61.90 & 62.93 & 40.82 & 55.00 & 65.63 \\
\bottomrule
\end{tabular}
}

\resizebox{\textwidth}{!}{
\begin{tabular}{@{}l c cc c cccc c@{}}
\toprule
\multirow{3}{*}{\textbf{Models}} & \textbf{TallyQA} & \multicolumn{3}{c}{\textbf{SpatialMM-Obj}} & \multicolumn{5}{c}{\textbf{3DSR-Bench-real}} \\
\cmidrule(lr){2-2} \cmidrule(lr){3-5} \cmidrule(lr){6-10}
& {\textbf{Avg.}} & {\textbf{1-obj}} & {\textbf{2-obj}} & {\textbf{Avg.}} & {\textbf{H.}} & {\textbf{L.}} & {\textbf{O.}} & {\textbf{M.}} & {\textbf{Avg.}} \\
\midrule
LLaVA-v1.5-7B & 58.74 & 57.37 & 44.87 & 48.91 & 55.42 & 57.82 & 26.09 & 39.42 & 45.02 \\
Prism-SigLIP-7B & 62.25 & 62.54 & 46.77 & 51.86 & 52.28 & 60.22 & 27.23 & 42.17 & 46.55 \\
Prism-DinoSigLIP-7B & 62.93 & 58.56 & 47.72 & 51.22 & 56.85 & 59.42 & 27.23 & 38.97 & 45.82 \\
\midrule
\rowcolor{blue!10}
\textbf{DyVA-7B} & 59.47 & 54.78 & 46.29 & 49.03 & 53.71 & 57.60 & 27.23 & 40.80 & 45.41 \\
\rowcolor{blue!10}
\textbf{DyVA-Qwen2.5-7B} & \bfseries 68.11 & 62.74 & 47.53 & \bfseries 52.44 & 52.57 & 54.51 & 27.23 &  49.60 & \bfseries 47.16 \\
\bottomrule
\end{tabular}
}
\end{table*}

Tab.~\ref{tab:sat_mmsi_mindcube} and \ref{tab:single_exp} present representative results under both single and multi-image settings. 
This framing allows us to disentangle how world-model features contribute across different reasoning regimes.

As presented in Tab.~\ref{tab:sat_mmsi_mindcube} and ~\ref{tab:single_exp}, we evaluate the OOD performance of DyVA-7B and DyVA-Qwen-2.5-7B. We examine DyVA’s performance relative to existing vision-language models across various spatial reasoning tasks. The key differences lie in DyVA’s use of Generative Encoders versus baselines that use only standard visual embeddings. Below, we discuss the strengths and weaknesses of DyVA in each benchmark category, drawing on the reported results of these tasks and models.

\textbf{DyVA can enable single-image trained WorldLMs to perform multi-image tasks exceptionally well.}
As in Tab.~\ref{tab:sat_mmsi_mindcube}, our best variant can perform strongly in multi-frame spatial understanding tasks. 

Specifically, on the MindCube benchmark (Tab.~\ref{tab:sat_mmsi_mindcube}), DyVA-Qwen2.5-7B achieves a new state-of-the-art performance with the highest overall score (49.8\% vs. 47.4\% for the runner-up baseline). It particularly excels in “Around” (rotating viewpoint) tasks (51.7\% vs. 44.1\%) and matches or slightly exceeds baselines on “Rot” tasks (37\% vs. ~36\%). These results suggest that DyVA latents significantly aid in tasks requiring mental rotation and perspective-taking, likely because they encode cross-view consistency. Specifically, these margins are consistent with the motion-consistency priors inherited from SVD’s pre-training on LVD-F video–text pairs~\cite{Blattmann2023Stable} that include how an object may appear from different angles.

This achievement is especially noteworthy considering the training efficiency. Compared to baselines where LLaVA-One-Vision is trained on 4M multi-frame images, Intern 2.5-VL is pretrained with 16.3M samples, including multi-image and video data, and Qwen-2.5-VL is also pre-trained with a variety of data comprising videos and multi-images. These baselines also have several complex methods for image preprocessing, such as patchifying~\citep{li2024llavaonevisioneasyvisualtask}, processing at different fps~\citep{bai2025qwen25vltechnicalreport}, and high-res processing~\citep{internvl25}. In contrast, we trained our DyVA model using only the most basic processing methods with minimal amount of data.

Our modest training budget and intuitive multi-image inference method suggest that world model latents strongly enhance the spatial understanding on multi-image benchmarks. We also believe that the fusion of SVD with SigLIP is a key factor that directly improves multi-image reasoning abilities.

\textbf{DyVA excels in handling spatial relations, counting and object queries, and 3D Scenes.} 
In Single-Image Spatial Reasoning, DyVA’s world-model features boost performance on tasks emphasizing geometric and relational spatial reasoning (orientation, adjacency, multi-object spatial layouts), reflecting improved 3D awareness.

\quad 1. Visual Spatial Relations (VSR): DyVA (SigLIP+SVD) achieves the highest average score (67.1\%) across VSR subtasks (topology, proximity, projection, direction, adjacency, orientation, unaligned), outperforming the SigLIP-only baselines (64.9–65.5\%) in Tab.~\ref{tab:single_exp}. In particular, DyVA significantly improves orientation reasoning (61.2\% vs ~55–49\% for baselines) and proximity/topology, suggesting it can better encode spatial layouts and object alignment. 


\quad 2. Counting and Object Queries (TallyQA, SpatialMM-Obj): On TallyQA (visual counting), DyVA-Qwen2.5 excels (68.1\% average), well above Prismatic baselines (62–63\%) and LLaVA (58.7\%)Tab.~\ref{tab:single_exp}. For the SpatialMM-Obj task (single- vs multi-object queries), DyVA-Qwen2.5 again slightly outperforms others (52.4\% vs ~51.8\% baseline) on the combined 1- and 2-object questions.

\quad 3. 3D Scene Reasoning (3DSR-Bench-real): This benchmark measures 3D spatial and depth understanding in real images. Notably, DyVA greatly improves the “Multiple objects” (M) subset (49.6\% vs ~40\% for baselines). This aligns with the conception that SVD latents capture implicit depth and occlusion cues learned from video modeling.

\textbf{Limitations and Areas for Improvement.} Despite its strengths in spatial reasoning, DyVA exhibits certain limitations, particularly on tasks that rely heavily on semantic language priors, non-canonical object arrangements, or temporal sequence understanding.

1. Weakened Performance on Language-Intensive Tasks: The fusion of world-model tokens can dilute the semantic precision required for certain tasks. As shown in Tab.~\ref{tab:encoder}, on benchmarks such as VQAv2 and TextVQA, which demand strong language priors and OCR capabilities, DyVA underperforms compared to SigLIP-only baselines. This suggests that while SVD latents enhance spatial awareness, they can interfere with fine-grained semantic grounding and text recognition, where the original visual features are more direct and precise.

2. Bias Towards Canonical Scene Structures: As previously noted in the VSR analysis, DyVA's performance drops significantly on the “Unaligned” subtask (57.5\% vs. 67.5\%). This indicates that embedding world-model context can be detrimental when objects lack canonical alignments. The model's latent prior appears biased toward common or expected scene structures, hindering its ability to reason about novel or unusual spatial configurations.

3. Less Reliable Sequential and Temporal Reasoning: The current SVD latents are less effective for understanding dynamic sequences. This is evidenced by a large performance drop in SAT Action Sequence and mixed results on MMSI. These outcomes suggest that the latents, while powerful for static scenes, are less reliable for predicting discrete action orders or interpreting rapid changes over time, marking a clear area for future improvement.

\section{Design-Space Exploration: Why DyVA Works}
\label{sec:designspace}

\textbf{Generative Encoders rely on both dynamic frames and text-aligned semantics as support.}

Building on the strong spatial performance demonstrated in both single-image and multi-image tasks in our experiments, we further analyze two key design axes to investigate the sources of WorldLM’s benefits: (i) the choice of different semantic vision encoders, and (ii) the potential of leveraging text-loss to supervise the joint-training of VAE and U-Net in SVD.

\begin{table*}[ht]

\centering
\caption{
    \textbf{Performance Comparison of SVD-based Vision Models.}
    \small Benchmark scores across a set of VQA, reasoning, and spatio-temporal tasks. All experiments use the LLaMA-2 7B backbone. The highest score in each column is marked in \textbf{bold}, and the second-highest is \underline{underlined}. Align: one-time alignment on LAION-558k~\cite{schuhmann2022laion5bopenlargescaledataset}. F1: one-time finetuning. Fused: 3-layer MLP projector.
}
\resizebox{\textwidth}{!}{%
\begin{tabular}{
    lcccccccccc
}
\toprule
\textbf{Model} & \textbf{Align} & {\textbf{VQAv2}} & {\textbf{GQA}} & {\textbf{VizWiz}} & {\textbf{VSR}} & {\textbf{POPE}} & {\textbf{TallyQA}} & {\textbf{SeedBench}} & {\textbf{SpatialMM}} & {\textbf{3DSR}} \\
\midrule
\multirow{2}{*}{\textbf{VAE-Only}}
    & $\times$ & 46.98 & 40.53 & 38.90 & 52.04 & 66.42 & 39.55 & 38.18 & 38.81 & 44.15 \\
    & $\checkmark$ & 50.70 & 43.26 & 48.67 & 52.29 & 60.80 & 42.48 & 41.53 & 37.3 & 43.43 \\
\midrule
\multirow{2}{*}{\textbf{SVD-Only}}
    & $\times$ & 63.51 & 55.18 & 44.95 & 57.93 & 82.38 & 49.75 & 50.15 & 42.03 & 42.93 \\
    & $\checkmark$ & 61.82 & 50.20 & 50.60 & 53.60 & 75.61 & 53.27 & 52.55 & 40.60 & 43.50 \\
\cdashline{2-11}
\addlinespace[3pt]
\quad \textbf{U-Net Trainable} & $\checkmark$ & 63.36 & 54.49 & 50.24 & 57.93 & 79.88 & 51.51 & 52.76 & 40.80 & 43.43 \\
\cdashline{2-11}
\addlinespace[3pt]
\quad \textbf{U-Net \& VAE Trainable} & $\checkmark$ & 60.99 & 49.80 & 50.17 & 52.53 & 77.08 & 53.75 & 52.33 & 39.50 & 44.00 \\
\midrule
\multirow{2}{*}{\textbf{Dino + SVD}}
    & $\times$ & 68.77 & 58.50 & 50.73 & 62.52 & 85.25 & 52.78 & 55.19 & 44.79 & 44.26 \\
    & $\checkmark$ & 68.44 & 55.57 & 51.13 & 59.41 & 85.54 & 54.15 & 56.49 & 43.40 & 45.07 \\
\midrule
\multirow{2}{*}{\textbf{SigLIP + SVD}}
    & $\times$ & \textbf{75.36} & \textbf{61.52} & \textbf{55.95} & \textbf{67.10} & 85.97 & \textbf{59.47} & \textbf{66.61} & \textbf{49.03} & 45.40 \\
    & $\checkmark$ & 73.63 & 58.89 & 54.63 & 61.62 & 84.37 & 56.98 & 62.09 & 45.40 & \underline{45.49} \\
\cdashline{2-11}
\addlinespace[3pt]
\quad \textbf{U-Net Trainable} & $\checkmark$ & 74.02 & 59.86 & 54.60 & 62.27 & 85.61 & 57.42 & 63.39 & 45.95 & 44.11 \\
\midrule
\multirow{2}{*}{\textbf{CLIP + SVD}}
    & $\times$ & 73.51 & 59.67 & 53.14 & 64.89 & 85.80 & \underline{58.25} & \underline{65.45} & 46.07 & \textbf{46.13} \\
    & $\checkmark$ & 72.99 & \underline{60.74} & \underline{55.89} & \underline{65.38} & 85.80 & 55.37 & 65.33 & 46.70 & 44.42 \\
\midrule
\multirow{2}{*}{\textbf{DinoSigLIP + SVD}}
    & $\times$ & \underline{74.28} & 60.16 & 54.13 & 64.81 & \textbf{87.27} & 57.42 & 64.54 & \underline{48.65} & 44.15 \\
    & $\checkmark$ & 72.42 & 59.28 & 54.47 & 61.29 & \underline{86.75} & 54.98 & 61.54 & 47.00 & 45.14 \\
\bottomrule
\end{tabular}
}
\label{tab:encoder}
\vspace{-1em}
\end{table*}

\subsection{Why do VAE, DINO, SVD-Only not work, but SigLIP+SVD does?}


To investigate the respective roles of the generative encoder and the semantic vision encoder within WorldLM, we conduct a two-stage ablation study. \textbf{First}, in a setting without the semantic vision encoder, we decouple the generative encoder into its component VAE and the complete generative encoder architecture. We then train and comparatively evaluate the performance of two distinct encoding approaches: one employing only the VAE for encoding and the other utilizing the entire generative encoder (SVD). \textbf{Second}, while keeping the generative encoder fixed, we systematically substitute the backbone of the semantic vision encoder with various alternative architectures to analyze its impact on the model's overall performance.
Our quantitative experimental results are presented in Tab.~\ref{tab:encoder}. 

\textbf{Prediction Matters.} The inference protocol for the SVD encoder is detailed in Sec.~\ref{sec:preliminary}. A similar inference process is employed when using VAE as the generative encoder. In contrast to extracting features from the layer before the middle block of U-Net, we directly use the features encoded by the VAE. To align the feature dimensionality with that of the SVD, we prepend several convolutional layers to the projector.
As evidenced by our experimental results in Tab.~\ref{tab:encoder}, the model employing only VAE for encoding exhibits a performance degradation across nearly all benchmarks when compared to models using SVD. This finding underscores the significance of the predicted dynamics for the WorldLM.

\textbf{WorldLMs need a text-aligned encoder.} Although SigLIP~\citep{zhai2023sigmoidlosslanguageimage} has recently shown dominant performance as an emerging vision encoder in current state-of-the-art VLMs, such as LLaVA-One-Vision~\citep{li2024llavaonevisioneasyvisualtask} and Prismatic-VLM~\citep{karamcheti2024prismaticvlmsinvestigatingdesign}, in this study, we investigate the respective roles of SigLIP, CLIP~\citep{Radford2021CLIP},  DINOv2~\citep{oquab2024dinov2learningrobustvisual}, and a combined DINO-SigLIP architecture as the semantic vision encoder. To ensure a fair comparison, we selected the ViT-L version for each model, all configured for a $224 \times 224$ input resolution. Furthermore, we adopted a consistent image processing strategy, which involves scaling and then cropping all images to uniform resolutions.

As demonstrated in Tab.~\ref{tab:encoder}, models that utilize SigLIP (including the DINO-SigLIP combination) or CLIP as the semantic vision encoder significantly outperform the model using DINOv2. Furthermore, when considering the aforementioned investigation of the generative encoder, the model with DINOv2 as the semantic vision encoder shows better performance than the generative-encoder-only architecture.

This leads to a key insight: for our WorldLM framework that is trained with text-loss supervision, in addition to predicted dynamic features, DyVA requires supplementary visual-semantic information from a model pre-trained on language-vision tasks (i.e., a text-aligned model). This insight also paves the way for future explorations: Can the generative encoder alone suffice to replace the semantic vision encoder? And is text-loss supervision the answer to WorldLM training?

\subsection{Can DyVA Benefit from U-Net \& VAE Training on Text-loss?}

We investigated the efficacy of fine-tuning the SVD's core components (U-Net and VAE) using only a text-loss signal. Our experimental results indicate this strategy is largely ineffective.

\textbf{Text supervision failed to help VQA tasks.} As shown in Tab.~\ref{tab:encoder}, making only the U-Net trainable yields inconsistent and marginal performance changes, while allowing both the U-Net and VAE to be trainable leads to a distinct and widespread degradation in performance across the benchmarks.

This suggests the high-level semantic supervision from the text-loss is ill-suited for adapting the low-level generative priors of these components. This constitutes one of the limitations of our current work. An alternative approach, inspired by methods like RAPE-E~\citep{leng2025repaeunlockingvaeendtoend}, involves aligning the features from the VAE and U-Net with the visual features from a semantic encoder such as DINOv2. Exploring such an alignment strategy is a promising direction for future research.

\section{Discussions and Outlooks}
\label{sec:discussion}

%

Over the recent past, video foundation models have demonstrated remarkable performance in key areas such as consistency and content generation. Through our empirical investigation on multimodal general tasks through a VLM framework, we find that:

\textbf{(1)} Paradigm comparisons reveal that WorldLM latents are powerful: these latents enable effective spatial and multi-view reasoning.

\textbf{(2)} Design-space explorations clarify which architectural choices benefit WorldLMs, while benchmark diagnostics explain where DyVA excels. 

\textbf{(3)} WorldLM encoders unlock visual reasoning through leveraging SVD’s predictive pre-training supplies transferable camera-motion and interaction priors, yet semantic gaps persist until the generative encoders are co-trained or better aligned with language signals.

Overall, we observe that WorldLM encoders offer a reliable pathway to stronger spatial and multi-view reasoning, and scaling trends in video generation~\citep{wiedemer2025videomodelszeroshotlearners,chi2025wowworldomniscientworld} suggest the semantic deficit may narrow as these models progress. Closing that gap for VLMs will likely require tighter alignment between dynamics-rich latents and language-grounded objectives.

\textbf{Outlooks.} Promising next steps include: (i) exploring text-to-video generators as encoders to test whether text-aligned priors further boost visual understanding; and (ii) designing specialized training that aligns generative latents with semantics without eroding their physical fidelity.


\bibliography{iclr2026_conference}
\bibliographystyle{iclr2026_conference}

\newpage
\appendix

\section*{Outline}

\begin{itemize}
    \item \textbf{Related Work (Section~\ref{app:relatedwork}):} Reviews prior work in three key areas: (1) predictive World Models, (2) diffusion-based Generalist Models for in-context learning, and (3) the application of diffusion models to discriminative vision tasks.

    \item \textbf{Model Formalization:} Details our architecture, including:
    \begin{itemize}
        \item Static visual features from a SigLIP encoder (Eq.~\ref{eq:proj-siglip}).
        \item Dynamic features from SVD U-Net hidden states (Eq.~\ref{eq:euler}, \ref{eq:unet-hidden}).
        \item The fusion mechanism for static and dynamic tokens (Eq.~\ref{eq:proj-svd}).
    \end{itemize}

    \item \textbf{Training Hyperparameters:} Specifies all training configurations, which are listed in Table~\ref{tab:training_hyperparameters}.

    \item \textbf{Design Space Explorations:} Presents key ablation studies, demonstrating:
    \begin{itemize}
        \item The model's sensitivity to temporal frames over spatial resolution (Table~\ref{tab:model_performance}).
        \item The rationale for our SVD feature fusion strategy, with comparative results in Table~\ref{tab:svd-middleblock}.
    \end{itemize}
\end{itemize}

\section{Related Work}

\label{app:relatedwork}

\subsection{World Models} Various methods have been developed to learn predictive models of visual dynamics. \cite{Ha2018WorldModels} proposed the original World Models framework, which learns a compressed latent representation of an environment’s dynamics using generative RNNs~\citep{Ha2018WorldModels}. Hafner introduced PlaNet \citep{Hafner2018PlaNet} and later Dreamer \citep{Hafner2019Dreamer}, which use latent space dynamics models trained on pixel observations for planning and control. More recently, large-scale self-supervised video models have emerged. For example, Stability AI’s Stable Video Diffusion trains a high-capacity latent video diffusion model on vast video datasets for high-quality text-to-video and image-to-video generation \citep{Blattmann2023Stable}.  Zhou (2024) introduced DINO-WM, a world model that leverages pretrained DINOv2 patch features to enable zero-shot goal-reaching via planning in feature space \citep{Zhou2024DINOWM}. Meta’s V-JEPA 2 \citep{Assran2025VJEPA2} and NVIDIA’s Cosmos platform \citep{Agarwal2025COSMOS} provide video foundation models that enable understanding, prediction, and planning from raw visual data.

\subsection{Generalist Models} Recent work has explored using diffusion-based generative models for flexible multi-task and in-context learning. \citet{Wang2023PromptDiffusion} presented Prompt Diffusion, a method that enables in-context learning in diffusion models by conditioning on example input-output image pairs and a text prompt. \citet{Geng2023InstructDiffusion} proposed InstructDiffusion, a unified framework that casts diverse vision tasks as a pixel-space image manipulation guided by human instructions, learned via a diffusion process. \citet{Bai2024Sequential} introduced a sequential modeling approach that represents images and annotations as “visual sentences,” enabling training a single large vision model across many tasks without using any language data. \citet{Lin2025RealGeneral} presented RealGeneral, which reformulates image generation as conditional frame prediction analogous to LLM in-context learning: using video diffusion models with novel modules, they unify multiple image-generation tasks (e.g., custom generation, canny-to-image) within one framework. Recently, Bagel~\citep{deng2025emergingpropertiesunifiedmultimodal} further extends these ideas by introducing novel techniques for improving the generalization and efficiency of multi-task learning in diffusion models.

\subsection{Diffusion models on Vision Tasks}

Recently, diffusion models, having established state-of-the-art performance in image generation, are increasingly being explored for their potential in discriminative vision tasks. This trend continues the historical trajectory to leverage generative models for discriminative tasks~\citep{HINTON2007535}. Current research has primarily followed three strategies for repurposing these models. The first utilizes them as potent feature extractors, leveraging the rich internal representations from frozen, large-scale text-to-image models for tasks like open-vocabulary panoptic segmentation \citep{baranchuk2022labelefficientsemanticsegmentationdiffusion, xu2023openvocabularypanopticsegmentationtexttoimage}. The second employs them at inference time as probabilistic world models, providing generative feedback to adapt discriminative models \citep{prabhudesai2023diffusionttatesttimeadaptationdiscriminative}. A third strategy directly leverages the model's likelihood estimation capabilities, reframing classification as an "analysis-by-synthesis" problem to perform zero-shot classification without additional training \citep{li2023diffusionmodelsecretlyzeroshot}.
More recently, a fundamental paradigm shift has emerged, reformulating core discriminative tasks as conditional denoising problems. This moves beyond using diffusion models as auxiliary tools, making the generative process itself the core mechanism for prediction. Seminal works in this area include DiffusionDet \citep{chen2022diffusiondet}, which frames object detection as a "noise-to-box" process of refining random boxes into precise detections, and DiffusionInst \citep{gu2022diffusioninstdiffusionmodelinstance}, which formulates instance segmentation as a "noise-to-filter" denoising process. This unified "denoising-as-prediction" framework replaces task-specific architectures (e.g., RPNs, query-based heads) with a single generative principle, marking a significant convergence and evolution in the modeling of discriminative vision tasks.


\section{Model Formalization}
\paragraph{VLM basics.} 
A frozen SigLIP image encoder $E_{\text{siglip}}$ maps an image 
$x \in \mathbb{R}^{H \times W \times 3}$ 
to a grid of patch embeddings 
$S \in \mathbb{R}^{N \times C_s}$, 
where $N$ is the number of patches and $C_s$ the channel width.  
A lightweight projector 
$P_{\text{siglip}} : \mathbb{R}^{C_s} \rightarrow \mathbb{R}^d$ 
aligns these to the LLM token space:
\begin{equation}
\label{eq:proj-siglip}
V_s = P_{\text{siglip}}(S) = \mathrm{MLP}_s(S) \in \mathbb{R}^{N \times d},
\end{equation}
where $\mathrm{MLP}_s$ is a 3-layer MLP with GELU activations.

\paragraph{SVD for single-image $\rightarrow$ video.}
Stable Video Diffusion (SVD) consists of a VAE encoder $\phi$ and a U-Net denoiser $f_\theta$ operating over a continuous noise scale $\sigma$ (Karras et al.).  
Given a conditioning image $x$, we compute a latent $z_0 = \phi(x)$.  
To form a video latent tensor, we replicate $z_0$ across $T$ frames:
\[
Z_0 = [z_0,\ldots,z_0] \in \mathbb{R}^{T \times C \times H' \times W'}.
\]

Let $\sigma_0$ denote the initial noise level from the SVD schedule.  
We perform one explicit Euler integration step over the ODE at $\sigma_0$ (classifier-free guidance disabled):
\begin{equation}
\label{eq:euler}
Z_{1} = Z_{0} + \Delta\sigma \, f_\theta(Z_{0}, \sigma_0, c),
\end{equation}
where $c$ denotes SVD conditioning (e.g., time/frame embeddings, text/image prompts), and $\Delta\sigma$ is the step size.  

We do not render frames; instead, we extract a U-Net hidden activation at the lowest spatial resolution on the downsampling path before the mid-block:
\begin{equation}
\label{eq:unet-hidden}
H \in \mathbb{R}^{T \times H_d \times W_d \times C_h} 
= \mathrm{Hidden}^{\text{pre-mid}}(f_\theta, Z_{1}).
\end{equation}

\paragraph{Multi-image extension.}
For multiple images $\{x_k\}_{k=1}^K$, we first compute their latents $\{z^{(k)}_0\}$.  
These are inserted as keyframes within $T$ frames at indices 
$i_k = \mathrm{round}(\mathrm{linspace}(0, T{-}1, K))$.  
We initialize $Z_0$ with copies of $z^{(1)}_0$ and set 
$(Z_0)_{i_k} \leftarrow z^{(k)}_0$ before the Euler step, yielding multi-image-aware $H$.

\paragraph{Static+dynamics token fusion.}
We convert $H$ into a token sequence by flattening spatial locations: 
$L = H_d W_d$, 
$\tilde{H} \in \mathbb{R}^{(T \cdot L) \times C_h}$.  
A projector 
$P_{\text{svd}} : \mathbb{R}^{C_h} \rightarrow \mathbb{R}^d$ 
maps these to the LLM token space:
\begin{equation}
\label{eq:proj-svd}
V_d = P_{\text{svd}}(\tilde{H}) = \mathrm{MLP}_d(\tilde{H}) \in \mathbb{R}^{M \times d},
\end{equation}
where $M = T \cdot L$.  

The SigLIP tokens $\hat V_s$ (Eq.~\ref{eq:proj-siglip}) are concatenated with $\hat V_d$ to form the visual sequence:
\[
V = [\hat V_s; \hat V_d].
\].


\section{Training Hyperparameters}

We adopt the hyperparameters in Table~\ref{tab:training_hyperparameters} for all our models (for both DyVA-7B and DyVA-Qwen2.5-7B).

\begin{table}[htbp]
\centering
\caption{Training Hyperparameters}
\label{tab:training_hyperparameters}
\begin{tabular}{ll}
\toprule
\textbf{Hyperparameter} & \textbf{Value} \\
\midrule
Batch Size & 128 \\
Max Gradient Norm & 1.0 \\
Weight Decay & 0.1 \\
Learning Rate & 2e-5 \\
Optimizer & AdamW \\
Scheduler & Warmup \& Cosine Decay \\
Warmup Ratio & 0.03 \\
\bottomrule
\end{tabular}
\end{table}

\section{More Design Space Explorations}

\textbf{WorldLM is sensitive to temporal information but demonstrates robustness to spatial resolution.} This dual characteristic is evident from our ablation studies. First, as detailed in Table \ref{tab:model_performance}, increasing the number of input frames from 1 to 14 yields a consistent and significant improvement across most benchmarks, such as VQAv2 (59.38 to 61.73). This highlights the model's proficiency in leveraging richer temporal context. Conversely, the impact of spatial resolution appears marginal. By comparing the results in Table \ref{tab:model_performance} (at 576×1024 resolution) with those in Table \ref{tab:encoder}, we find that variations in resolution do not lead to substantial performance changes. These combined findings suggest that our model architecture prioritizes temporal patterns over high-frequency spatial details for the evaluated tasks.

\begin{table*}[htbp] 
\centering 
\caption{Model Performance Across Different Frame Numbers. These are DyVA with SVD only encoders using a image resolution of $576\times1024$} 
\label{tab:model_performance} 

\resizebox{\textwidth}{!}{%
\begin{tabular}{l cc ccc ccc ccc}
\toprule
\textbf{Frames} & \textbf{Pretrain} & \textbf{Tuning} & \textbf{VQAv2} & \textbf{GQA} & \textbf{VizWiz} & \textbf{VSR} & \textbf{POPE} & \textbf{TallyQA} & \textbf{SeedBench} & \textbf{SpatialMM-Obj} & \textbf{3DSR-Bench-real} \\
\midrule
1  & 558k & 665k & 59.38 & 47.75 & 48.74 & 52.12 & 75.74 & 50.97 & 51.12 & 38.81 & 45.40 \\
4  & 558k & 665k & 60.10 & 47.36 & 46.24 & 53.19 & 77.60 & 50.68 & 52.24 & 42.48 & 45.67 \\
8  & 558k & 665k & 60.80 & 48.63 & 50.25 & 52.20 & 78.15 & 51.46 & 52.81 & 37.98 & 46.32 \\
14 & 558k & 665k & 61.73 & 49.71 & 38.68 & 53.43 & 78.80 & 52.19 & 53.28 & 39.78 & 46.32 \\
\bottomrule
\end{tabular}%
}
\end{table*}

\textbf{Fusing SVD latents after the U-Net's middle block substantially improves performance over the baseline fusion strategy.} We further explore the optimal strategy for integrating SVD-derived temporal latents into the model architecture. Specifically, we compare our baseline DyVA-SVD model with a variant, DyVA-SVD-Post-MiddleBlock, which injects the latents after the U-Net's middle block. The results, presented in Table \ref{tab:svd-middleblock}, indicate that the Post-MiddleBlock fusion strategy yields significant performance gains across most benchmarks. Notably, we observe substantial improvements on GQA (+4.1), VSR (+4.09), and POPE (+4.56), strongly advocating for this modified fusion approach and highlighting the critical impact of architectural choices in temporal feature integration. However, given the absence of across-the-board performance gains, and in consideration of inference efficiency, we ultimately adopted the "pre-mid" implementation.

\begin{table}[h]
\centering
\caption{\textbf{SVD vs. SVD-MiddleBlock.} Comparison of different fusion strategies using SVD latents.}
\scriptsize
\setlength{\tabcolsep}{3pt}
\begin{tabular}{lcccccccccc}
\toprule
\textbf{Model} & \textbf{VQAv2} & \textbf{GQA} & \textbf{VizWiz} & \textbf{VSR} & \textbf{POPE} & \textbf{TallyQA} & \textbf{SeedBench} & \textbf{Spatial} & \textbf{3DSR} \\
\midrule
\textbf{SVD-Only}             & 61.82 & 50.20 & 50.60 & 53.60 & 75.61 & 53.27 & 52.55 & 40.60 & 43.50 \\
\textbf{SVD-Only-Post-MiddleBlock}  & 62.86 & 54.30 & 51.41 & 57.69 & 80.17 & 51.36 & 52.50 & 41.13 & 43.84 \\
\bottomrule
\end{tabular}
\label{tab:svd-middleblock}
\end{table}

\end{document}